\def\year{2022}\relax
\documentclass[letterpaper]{article} 
\usepackage{aaai22}  
\usepackage{times}  
\usepackage{helvet}  
\usepackage{courier}  
\usepackage[hyphens]{url}  
\usepackage{graphicx} 
 \usepackage{multirow}
  \usepackage{multicol}
\usepackage{natbib}  
\usepackage{caption} 
\DeclareCaptionStyle{ruled}{labelfont=normalfont,labelsep=colon,strut=off} 
\usepackage{algorithm}
\usepackage{algorithmic}
\usepackage{booktabs}
\usepackage{epsfig}
\usepackage{xcolor}
\newcommand{\mldam}{$DA^{3}$\space}

\usepackage{amsmath,amsfonts,bm}

\def\eqref#1{equation~\ref{#1}}

\def\1{\bm{1}}

\DeclareMathAlphabet{\mathsfit}{\encodingdefault}{\sfdefault}{m}{sl}
\SetMathAlphabet{\mathsfit}{bold}{\encodingdefault}{\sfdefault}{bx}{n}
\newcommand{\tens}[1]{\bm{\mathsfit{#1}}}
\def\tA{{\tens{A}}}

\def\tG{{\tens{G}}}
\def\tH{{\tens{H}}}

\def\tM{{\tens{M}}}

\def\tW{{\tens{W}}}

\def\gG{{\mathcal{G}}}

\newcommand{\R}{\mathbb{R}}

\usepackage{newfloat}
\usepackage{listings}

\floatstyle{ruled}
\newfloat{listing}{tb}{lst}{}
\floatname{listing}{Listing}
\title{$\textbf{DA}^{\textbf{3}}$: \underline{D}ynamic \underline{A}dditive \underline{A}ttention \underline{A}daption for Memory-Efficient \\ On-Device Multi-Domain Learning}

\author{Li Yang, Adnan Siraj Rakin and Deliang Fan}
\affiliations{
School of Electrical, Computer and Energy Engineering, Arizona State University, Tempe, AZ 85287\\
{\tt\small  lyang166@asu.edu, asrakin@asu.edu, dfan@asu.edu}}

\usepackage{bibentry}

\begin{document}

\maketitle
\vspace{-1em}
\begin{abstract}
Nowadays, one practical limitation of deep neural network (DNN) is its high degree of specialization to a single task or domain (e.g., one visual domain). It motivates researchers to develop algorithms that can adapt DNN model to multiple domains sequentially, while still performing well on the past domains, which is known as multi-domain learning. Almost all conventional methods only focus on improving accuracy with minimal parameter update, while ignoring high computing and memory cost during training, which makes it difficult to deploy multi-domain learning into more and more widely used resource-limited edge devices, like mobile phone, IoT, embedded system, etc. During our study in multi-domain training process, we observe that large memory used for activation storage is the bottleneck that largely limits the training time and cost on edge devices. To reduce training memory usage, while keeping the domain adaption accuracy performance, we propose \textbf{Dynamic Additive Attention Adaption (\mldam)}, a novel memory-efficient on-device multi-domain learning method.
\mldam learns a novel \textit{additive attention adaptor} module, while freezing the weights of the pre-trained backbone model for each domain. Differentiating from prior works, such module not only mitigates activation memory buffering for reducing memory usage during training, but also serves as dynamic gating mechanism to reduce the computation cost for fast inference.
We validate \mldam on multiple datasets against state-of-the-art methods, which shows great improvement in both accuracy and training time. Moreover, we deployed \mldam into the popular NIVDIA Jetson Nano edge GPU, where the measured experimental results show our proposed \mldam reduces the on-device training memory consumption by \textbf{19-37}\textbf{$\times$}, and training time by \textbf{2}\textbf{$\times$}, in comparison to the baseline methods (e.g., standard fine-tuning, Parallel and Series Res. adaptor, and Piggyback).
\end{abstract}

\section{Introduction}

\begin{figure}[t]
  \centering
  \includegraphics[width=0.95\linewidth]{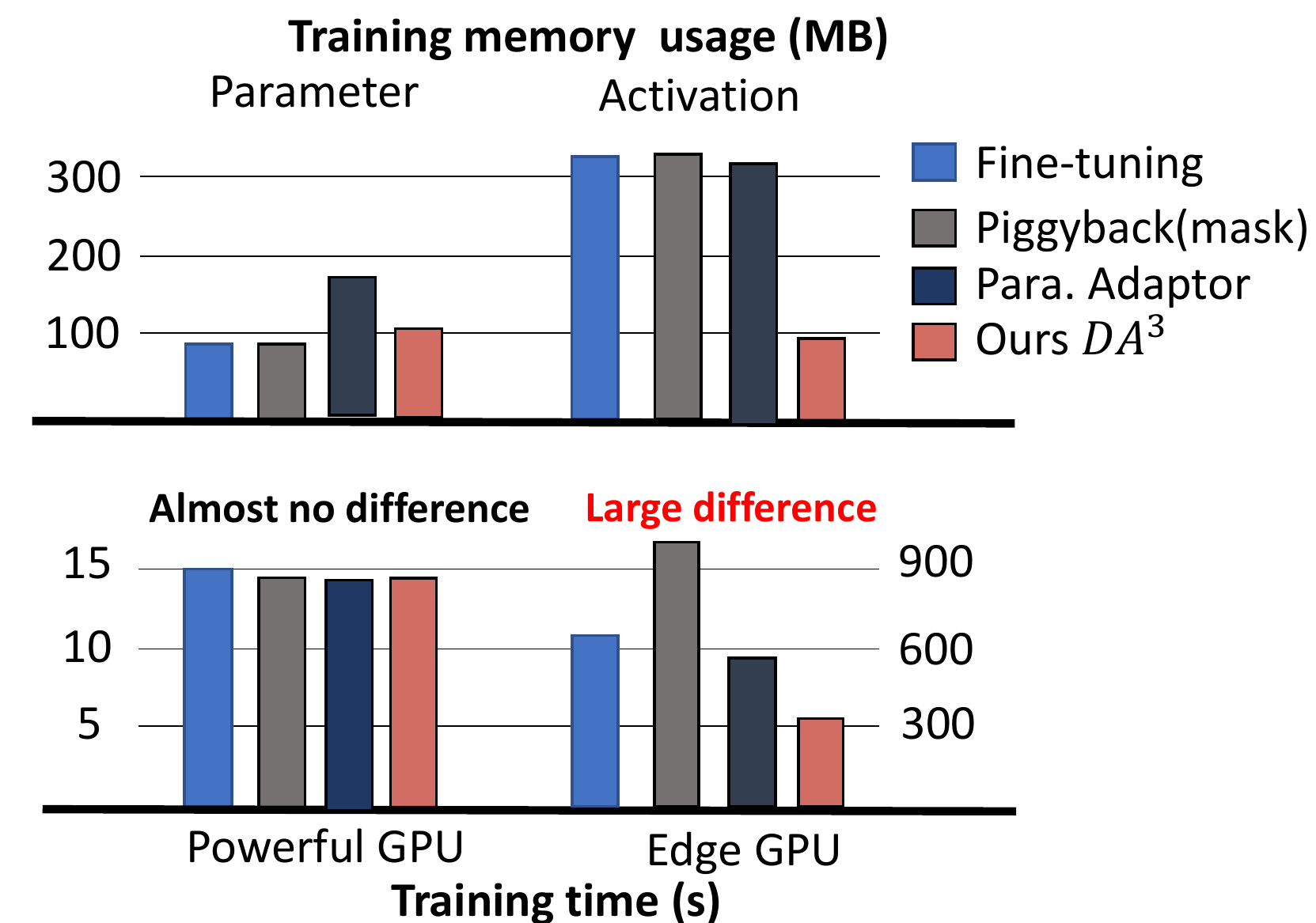}\\
\caption{An example of adapting ResNet50 (pre-trained on ImageNet dataset) to Flower dataset~\cite{nilsback2008automated}. \textit{Top:} model parameters and activation memory of three different methods. \textit{Bottom:} training time of one epoch on two different platforms: one powerful GPU (Quadro RTX 5000) and one edge GPU (Jetson Nano)
\label{fig:cover_page}
}
\label{fig:ch_att}
\end{figure}

Nowadays, one practical limitation of deep neural network (DNN) is its high degree of specialization to a single task or domain (e.g., one visual domain). It motivates researchers to develop algorithms that can adapt DNN model to multiple domains sequentially, while still performing well on the past domains. This process of gradually adapting DNN model to learn from different domain inputs over time is known as \textit{multi-domain learning}.
Nowadays, the utilization of IoT devices is greatly increased (e.g., 250 billion microcontrollers in the world today\footnote{https://venturebeat.com/2020/01/11/why-tinyml-is-a-giant-opportunity/}), that collect massive new data crossing various domains/tasks in our daily life. To process the new data, a general way is to perform learning/training on cloud servers, and then transfer the learnt DNN model back to IoT/edge devices for inference only. However, such method (\textit{ learning-on-cloud and inference-on-device}) is inefficient or unacceptable due to the huge communication cost between cloud and IoT/edge devices, as well as data-privacy concern (e.g., sensitive health care application).
These challenges lead to a recently rising research direction about `on-device multi-domain learning'. 

%

Conventional multi-domain learning methods can be mainly divided into three directions: fine-tuning based method, adaptor-based method, and mask-based method. As the first approach, inspired by the success of transfer learning, fine-tuning~\cite{kornblith2019better,cui2018large} is a natural approach to optimize the whole pre-trained model from old domains to new target domains. However, the training cost is huge since all the parameters need to be updated,
and the overall size of parameters will increase linearly w.r.t the number of domains. One alternative method is to only fine-tune the batchnorm and last classifier, but suffering from limited domain adaption capacity \cite{mudrakarta2018k}. 
In the second approach, 
\cite{rebuffi2017learning, rebuffi2018efficient} propose an adapter-based method, which learns a domain-specific residual adaptor while freezing the pre-trained model. In addition, such method also needs to fine-tune the batchnorm layer of the pre-trained model to avoid domain shift.
Different from that, Piggyback \cite{mallya2018piggyback}, as a representative work of the third mask-based learning approach, proposes to only learn a binary element-wise mask ($\{0,1\}$) w.r.t all weights, while keeping the pre-trained model fixed.  
Although adaptor- and mask-based methods reduce the training cost by freezing the pre-trained model compared with fine-tuning based method, the ways of learning domain-specific parameters both are inefficient in terms of activation memory efficient, which is the bottleneck to enable on-device multi-domain learning.    


To investigate the possibility of on-device learning for multi-domain learning methods, we tested three representative methods on these three directions respectively, in both powerful GPU (Nvidia RTX5000 used in desktop or cloud server training) and edge GPU (Nvidia Jetson Nano GPU used in edge device training). The measured training memory usage and training time are shown in Fig.\ref{fig:cover_page}.  

\vspace{3mm}
\textbf{Observation 1.} \textit{training process is memory-intensive, where the \textbf{intermediate activation buffering in memory during back-propagation} is the bottleneck (at least 3X more than model itself as shown in Fig.\ref{fig:cover_page}), to limit the speed of on-edge-device learning.} 
\vspace{3mm}

During training, the memory usage for activation storage (defined as \textit{activation memory} in this work) is almost 3X larger than the model itself. Such large training memory is not an issue (assuming with the same training time) in a powerful GPU with large enough memory capacity. However, for memory-limited edge GPU typically used in edge device training, such large memory usage becomes the bottleneck to limit training speed, and correspondingly leading to significantly different training speeds across different training methods for the same network and dataset as shown in Fig.\ref{fig:cover_page}.
Almost all prior domain adaption schemes only emphasize improving accuracy with minimal parameter update, while ignoring the computing- and memory-intensive nature of their methods, which makes it in-efficient to deploy into resource-limited edge-based training devices, like mobile phones, embedded system, IoT, etc.


In this work, we propose \emph{Dynamic Additive Attention Adaption (\mldam)},
a new training scheme for memory-efficient on-device multi-domain learning (simplified as on-device learning in this work). Differentiating from prior works, our \mldam is designed to eliminate the storage of intermediate activation feature map (i.e., dominating memory usage during on-device learning) to greatly reduce overall memory usage. Furthermore, to improve the adaption accuracy performance, \mldam is embedded with a novel \textit{dynamic additive attention adaptor} module, which is not only designed to avoid activation buffering for memory saving during training, but also reduces the computation cost through a dynamic gating mechanism. In summary, our technical contributions include:
\begin{itemize}
     \item  First, we present a complete analysis of memory consumption during training to prove that activation memory buffering is the key memory bottleneck during on-device multi-domain learning. More importantly, based on this analysis, we further discover an important observation to guide our design: the complete activation map (i.e., dominating memory usage) needs to be stored for backward propagation during training if it has \textbf{multiplicative relationship} with learned parameters (i.e., weight, mask), while the \textbf{additive relationship} (i.e., bias) is activation free. 

    \item Following our memory usage analysis, we propose a novel training method, called \textit{Dynamic Additive Attention Adaption (\mldam)}, for memory-efficient on-device multi-domain learning. The main idea of \mldam is that it freezes the learned parameters which has a multiplicative relationship with input activation, and only updates the learnable parameters that have an additive relationship. Thus, there is no need to store the memory-dominating activation feature map during backward propagation. Moreover, to further enrich the adaption capacity, we propose a novel \textit{additive attention adaptor} module that not only follows the additive principle to eliminate dominating activation memory buffering, but also implements a dynamic gating mechanism to reduce inference computation complexity. Such adaptor can plug in and play on the popular backbone model architectures for memory-efficient multi-domain learning.
    
    \item We conduct extensive experiments of the proposed \mldam method comparing with prior competitive baselines. \mldam could achieve state-of-the-art accuracy on popular multi-domain adaption datasets. More importantly, unlike previous methods, we, for the first time, test the training efficiency on an edge GPU (i.e., NVIDIA Jetson Nano) to prove \mldam could greatly reduce the training cost (i.e., both memory and time) in real device. The experimental results show that \mldam reduces the on-device training memory consumption by \emph{19-37}\emph{$\times$} and actual training time by \emph{2}\emph{$\times$} in comparison to the baseline methods (e.g., standard fine-tuning~\cite{guo2019spottune}, Parallel and Series Res. adaptor~\cite{rebuffi2017learning,rebuffi2018efficient}, and Piggyback~\cite{mallya2018piggyback})
\end{itemize}

\vspace{-1em}
\section{Related Work}
\subsection{Multi-Domain Learning}

Multi-domain learning~\cite{rebuffi2017learning, rebuffi2018efficient, rosenfeld2018incremental,cui2018large,wang2019towards,mancini2018adding} aims to build a model, which can adapt a task into multiple domains without forgetting previous knowledge, meanwhile learning as few parameters as possible. Series Res. Adaptor~\cite{rebuffi2017learning} addresses this challenge by learning additional residual adaptor for each layer while freezing the original pre-trained model except batch norm layer. Based on the same idea, the authors further update the topology of the residual adapter to be parallel rather than serial~\cite{rebuffi2018efficient}, resulting in better performance. Furthermore, \cite{berriel2019budget} proposes Budget-ware Adaptor which aims to reduce the model parameters to enable efficient inference, but has no benefit for training. \cite{wang2019towards} achieves training-efficiency by reducing the training run-times for different tasks. However, the training procedure is memory-intensive, which is impractical for resource-limited on-device learning.
\cite{rosenfeld2018incremental} proposes to recombine the weights of the backbone model via controller modules in channel-wise. In short, all these methods above tackle the multi-domain challenge by learning additional domain-specific modules. Different from that, Piggyback~\cite{mallya2018piggyback} learns task-specific binary masks for each task. It achieves this by first generating the real-value masks which own the same size with weights, then passing through a binarization function to obtain binary masks that are then applied to existing weights. Furthermore, \cite{mancini2018adding,yang2020ksm} combine the binary mask with additional reparametrization methods to increase adaption capacity, but suffering from even more computation and memory cost during the training procedure.

\subsection{Memory-Efficient Training}
There are two conventional techniques to reduce the activation memory: gradient checkpointing~\cite{chen2016training} and reversible network~\cite{gomez2017reversible}. Gradient checking only stores a subset of the network activations instead of all. As a consequence, the activations that are not stored must be recomputed during the backward pass. Furthermore, reversible network achieves backpropagation without storing activations, however, that means each layer's activations have to be recomputed exactly from that of the next layer. These two methods reduce the activation memory usage by involving additional computation. As mentioned in \cite{sohoni2019low}, for a 30\% increase in computing overhead, checkpointing can reduce the memory required for the activations by 5.8x. \cite{gomez2017reversible} also shows that the reversible netwo

rk has roughly 50\% computational overhead than ordinary backpropagation in practice. More recently, TinyTL \cite{cai2020tiny} proposes a lite residual learning module to lower the activation memory in transfer learning. However, this method involves additional training and searching procedure to find the suitable sub-network architecture for each target task. 


\subsection{Attention and Dynamic Mechanism}
Attention mechanism has proven to be a promising method to enhance DNN accuracy. SE-Net~\cite{hu2018squeeze} is the first that presents an effective mechanism to learn channel attention and achieves good performance. Later, CBAM~\cite{woo2018cbam} combines channel and spatial attention to enhance feature aggregation.
Furthermore, inspired by the attention mechanism, dynamic pruning aims to reduce the computation cost for fast inference. 
\cite{verelst2020dynamic} proposes a spatial dynamic convolution method, which adapts a residual block where a small gating branch learns which spatial positions are evaluated.
All of the above methods focus on developing sophisticated attention modules for better performance of a single task. They optimize these attention modules jointly with the original model. Different from them, we aim to utilize the attention mechanism to enhance the adaption capacity of the pre-trained model on multi-domain learning. 

\section{Memory Analysis in Multi-Domain Training}
\label{sec:rethinking}

In this section, we first explore the training memory usage under different multi-domain learning methods. Then, we will conduct a quantitative analysis of memory usage for each layer of DNN model. Moreover, such analysis will guide us to investigate a possible solution to achieve on-device memory-efficient learning method.

\paragraph{Fine-tuning and adaptor-based methods} Both Fine-tuning and adaptor-based training schemes are popular in this research area, which requires fine-tuning all or part of parameters in the pre-trained model. Fine-tuning based training method on the target dataset domain is intuitive to understand. But, to explain the adaptor-based method, we illustrate the architecture of two popular adaptor-based methods. Such method needs to fine-tune the additional convolution layer and the original batchnorm (BN) layers. 
To understand the training memory consumption, let's assume a linear layer whose forward process be modeled as: $a_{i+1} = a_i\tW + b$, then its back-propagation process is
\begin{equation}
\label{eq:act_loss}
\begin{gathered}
    \frac{\partial \mathcal{L}}{\partial a_i} = \frac{\partial \mathcal{L}}{\partial a_{i+1}} \frac{\partial a_{i+1}}{\partial a_{i}} = \frac{\partial \mathcal{L}}{\partial a_{i+1}} \tW^T, \\
    \frac{\partial \mathcal{L}}{\partial \tW} = \textcolor{red}{a_{i}} \frac{\partial \mathcal{L}}{\partial a_{i+1}}, 
    \frac{\partial \mathcal{L}}{\partial b} = \frac{\partial \mathcal{L}}{\partial a_{i+1}}
\end{gathered}
\end{equation}
According to Eq.~\ref{eq:act_loss}, to conduct conventional back propagation based training for entire model, model weights-$\tW$, gradients and activation-$a_i$ all need to be stored for computing, leading to large memory usage. However, it is interesting to see that, if only updating bias, which has an additive relationship with activation-$a_i$, no activation storage is needed since previous activation $a_i$ is not involved in the backward computation. The same phenomena can also be found in both Conv and BN layers.


\paragraph{Mask-based learning method}
For the mask-based learning method, assuming a linear layer whose forward process is given as: $a_{i+1} = a_i(\tW \cdot \tM)+b $, where $\tM$ is the mask to be learned with the same size as $\tW$. The weights-$\tW$ is fixed, while only training the mask-$\tM$. Then the backward process can be shown as:
\begin{equation}
\begin{gathered}
    \frac{\partial \mathcal{L}}{\partial \tM} = \textcolor{red}{a_{i}} \frac{\partial \mathcal{L}}{\partial a_{i+1}} \cdot \tW,
    \frac{\partial \mathcal{L}}{\partial b} = \frac{\partial \mathcal{L}}{\partial a_{i+1}}
\label{eqt:mask_loss}
\end{gathered}
\end{equation}
Eq.\ref{eqt:mask_loss} shows that learning mask needs to store not only activation-$a_i$, but also the mask-$\tW$ and weights-$\tW$ during training. In terms of computation, comparing Eq.\ref{eqt:mask_loss} with Eq.\ref{eq:act_loss}, such method also needs additional multiplication computation in both forward and backward pass. These observations explain why Piggyback has the largest training time in edge GPU as shown in Fig.\ref{fig:cover_page}.
Other mask-based methods~\cite{mancini2018adding, yang2020ksm} even need more computation cost than Piggyback, since they involve additional reparameterization techniques. In addition, similar to fine-tuning and adaptor-based methods, training bias does not involve activation storage.

\paragraph{Training memory usage analysis.}
\vspace{-1em}
\begin{table}[h]
\centering
\caption{Summary of the parameters and activation memory consumption of different layers. We denote the weights $\tW^{(l)} \in \R^{c_\textrm{in}\times c_\textrm{out}\times kh\times kw}$, where $c_\textrm{in}, c_\textrm{out}, kh, kw$ refers the weight dimension of $l$-th layer, including \#output channel, \#input channel, kernel height and width, respectively. We also denote the input activation $\tA^{(l)} \in \R^{n\times c_\textrm{in}\times h\times w}$, where $n, h, k$ refers the batch size, activation height and width, respectively.}
\vspace{-1em}
\scalebox{0.9}{
\begin{tabular}{@{}ccc@{}}
\toprule
Layer Type & Trainable Param. ($p$) & Activation ($a$) \\
Conv & $c_\textrm{in}\times c_\textrm{out}\times kh\times kw$ &  $n\times c_\textrm{in}\times h\times w$ \\
FC & $c_\textrm{in}\times c_\textrm{out} + c_\textrm{out}$ & $n\times c_\textrm{in}\times h\times w$ \\
BN &  $2\times c_\textrm{out}$ & $n\times c_\textrm{in}\times h\times w$ \\
ReLU & 0 & $n\times c_\textrm{in}\times h\times w$ \\
Sigmoid & 0 & $n\times c_\textrm{in}\times h\times$ w\\\bottomrule
\end{tabular}}
\label{tab:traininig}
\end{table}

Here, we first define training memory usage that will be used in the rest of this paper. As displayed in Table \ref{tab:traininig}, memory usage is proportional to the number of parameters during training, which can be treated as two main groups: i) \# of trainable parameters - $p$ (i.e. weights, bias) and gradient of each parameter; ii) activation memory consisting of the feature maps stored to update the parameters of previous layers using the chain rule. Note, trainable parameter memory has the same size as gradient memory. We only list \# of trainable parameters - $p$ in Table \ref{tab:traininig}.

For most convolution layers, kernel height/width is much smaller than activation channel width/height (i.e., kh $\ll$ h; kw $\ll$ w). Thus, for a moderate batch size (e.g., $n = 64/128/256$), activation memory size is much larger than that of trainable parameters (i.e., $a \gg p$). More interestingly, even though BN and sigmoid function have a negligible amount of trainable parameters ($p$), both functions produce an activation output ($a$) of the same size as a CONV/FC layer. 

From the above analysis, it can be easily seen that \textbf{DNN training memory usage is dominated by the activation feature map storage rather than the model parameter itself}. It is important to optimize the activation feature map memory usage if targeting memory-efficient learning. As for existing multi-domain learning methods, both mask-based and fine-tuning methods require heavy memory consumption during the backward propagation, requiring all weights, gradients, activation storage. Moreover, extra mask memory is required for mask-based method.
It is also interesting to observe that, if it is possible to only update bias in multi-domain learning, the dominating memory usage component - activation is not required anymore. It is because bias has an additive-only relationship with input activation, enabling backward propagation independently.
Based on above analysis, we summarize the underlying reason as the observation-2 below, which motivates and justifies our proposed \mldam method in the next section.

\vspace{2mm}
\textbf{Observation 2.} \textit{The complete activation map needs to be stored for backward propagation during training if it has the \textbf{multiplicative relationship} with learned parameter (i.e., weight, mask), while the \textbf{additive relationship} (e.g., bias) is activation free.}

\begin{figure*}[h]
  \centering
  \includegraphics[width=0.75
  \linewidth]{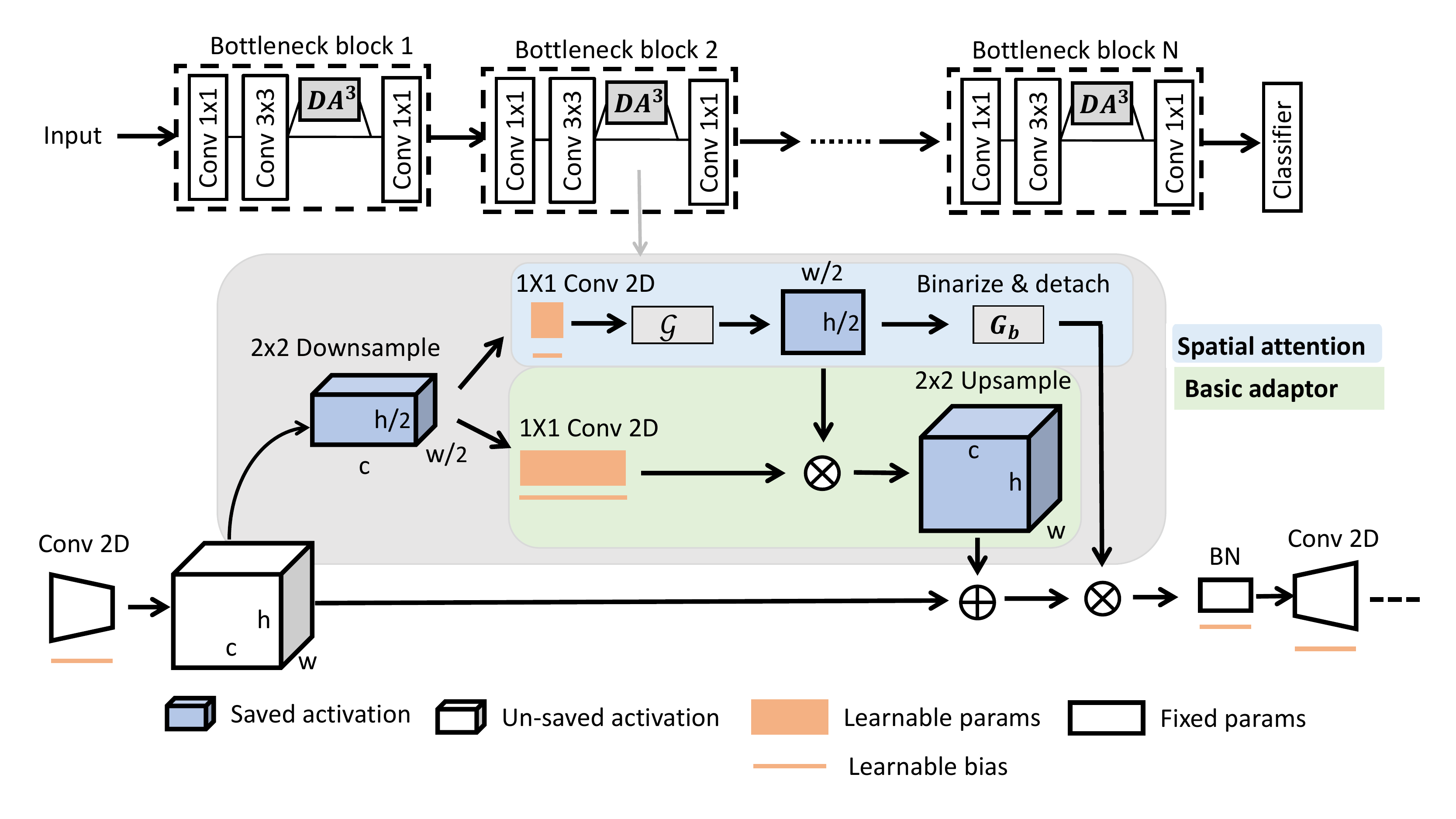}\\
\vspace{-1.5em}
\caption{Overview of the proposed \mldam, consisting of spatial attention (top branch) and basic adaptor (bottom branch).}
\label{fig:att_adapter}
\end{figure*}

\section{Proposed Method}
\label{sec:da2}

Motivated by the above memory usage analysis, we propose a new training method, named \textit{Dynamic Additive Attention Adaption} (\mldam) as illustrated in Fig.\ref{fig:att_adapter}. \mldam introduces a novel \textit{additive attention adaptor} module in each block for a given DNN model, that follows the additive relationship with the weight of the main branch (i.e., pre-trained model) as mentioned in \textbf{observation 2}. To learn each new domain, \mldam only updates the additive attention adaptor and the bias of the pre-trained model, while freezing the corresponding weight to preserve the knowledge of the previous domains.   
As the detailed structure of the additive attention adaptor illustrated in Fig.\ref{fig:att_adapter}, it aims to refine the activation of the pre-trained model, which is computed as:
\begin{equation}
    \tA{_i^*} = (\tA \textcolor{red}{+}    \tens{\tH}(\tA)) 
\label{eqt:basic}
\end{equation}
Where $\tH$ denotes the output activation of the additive attention adaptor module. To design an efficient yet powerful module, we first compute the spatial attention $\tH_s$ and the basic adaptor $\tH_a$ at two parallel branches, then combining them as:
\begin{equation}
    \tH(\tA) = \gG(\tH_s(\tA)) \otimes \tH_a(\tA)
\end{equation}
Where $\otimes$ denotes the element-wise multiplication and $\gG(\cdot)$ is a Gumbel-Softmax~\cite{jang2017categorical} function to obtain the spatial-wise soft attention of the basic adaptor activation $\tH_a$. 
Furthermore, instead of fully utilizing the pre-trained model as prior works~\cite{rebuffi2017learning, rebuffi2018efficient,cai2020tiny}, we select the important weights for current domain by 
turning the soft attention $\gG(\tH_s(\tA))$ into binary hard gating $\tG_b \in \{0,1\}$.
Then, the Eq.\ref{eqt:basic} can be further modified as: 
\begin{equation}
    \tA{_i^*} = (\tA  \textcolor{red}{+} \tens{\tH}(\tA) ) \otimes \tG_{b}^{detach} 
\label{eqt:detach}
\end{equation}
Importantly, as the gating $\tG_b$ has the multiplicative relationship with the activation of the pre-trained model $\tA$, we detach the gating $\tG_b$ from backward computation graph. By doing so, the detached $\tG_{b}^{detach}$ is only used for forward pass that has no gradient to do backward propagation during training. Thus, it will not cause additional activation memory storage from the main branch pre-trained model.  

\textbf{Basic adapter branch.} 
As shown in Eq.\ref{eq:act_loss}, the activation size grows quadratically with the resolution (i.e., height and width). Thus, to reduce the activation size, in the basic adapter branch, a 2×2 average pooling is used to down-sample the input feature map, followed by a 1x1 convolution layer.  

\textbf{Spatial attention branch.}
To sample the activation in spatial-wise (i.e.,${n\times 1\times h/2\times w/2}$)  after down-sampling, we adopt a 1x1 convolution layer with the output channel as 1. Then, following the Gumbel-softmax function $\gG(\cdot)$, we obtain the soft attention. Such soft attention plays two roles: 1) it will be multiplied with the basic adaptor output to strengthen the domain-refined activation; 2) it turns to binary hard gating $\tG_b \in \{0,1\}$ by applying a binarization trick and then multiply with the output of main branch activation. By doing so, it could dynamically select the input-relevant spatial position for current domain. To avoid the activation storage of the main branch during training, the binary gating is detached from the computation graph that has no gradient to do backward propagation. The detailed Gumbel-softmax and binarization trick are relegated to the appendix-B.

Following the spatial attention branch and basic adaptor branch, the up-sampled and domain-refined activation will be added to the main branch (pre-trained backbone model) output activation. Note that, different from the conventional attention scheme, where the output directly multiplies the main branch output activation, we design our additive attention adaptor in a way to \textit{add} it to the main branch. The main benefit of doing so is the proposed additive attention adaptor module can be processed during backward independently, without creating a new backward pass as in the traditional multiplication based mechanism.
Therefore, the increased memory usage for the proposed additive attention adaptor is very limited, which will be discussed in the later experimental section.


\textbf{Dynamic Additive attention adaptor integration}.
Fig.\ref{fig:att_adapter} illustrates an example to integrate the proposed additive attention adaptor in bottleneck block on ResNets\cite{he2016deep}. For the basic block which has two connected convolution layers, we plug in the additive attention adaptor after the last convolution layer. For the bottleneck block, 
the last convolution layer will enlarge the output channels (i.e. $4\times$), which increases the output activation linearly. To avoid involving large activation increases, we add the additive attention adaptor after the second convolution layer.


\section{Experiments}
\subsection{Experimental Setup}
\paragraph{Datasets and Evaluation Metrics.}

To evaluate the efficacy of the proposed \mldam method, we use standard and popular multi-domain learning dataset similar as many prior works \cite{mallya2018piggyback,guo2019spottune}. This setting includes five datasets (e.g., WikiArt \cite{saleh2015large}, Sketch \cite{eitz2012humans}, Standford Cars \cite{krause20133d}, CUBS\cite{wah2011caltech} and Flowers\cite{nilsback2008automated}). For each of the dataset, we report the test accuracy (\%) on the publicly available test set.

\begin{table*}[ht]
\centering
\caption{Summary of the results (i.e., accuracy \% ) of the proposed method and comparison with the baseline techniques on five datasets ( e.g., CUBS, Stanford Cars, Flowers, WikiArt and Sketches). }
\vspace{-1em}
\scalebox{0.9}{
\begin{tabular}{@{}ccccccc@{}}
\toprule
Model & CUBS & Stanford Cars & Flowers & WikiArt & Sketches & Average \\\midrule
Standard Fine-tuning\cite{guo2019spottune} & 81.86 & 89.74 & 93.67 & {75.60} & 79.58 & 84.09 \\
BN Fine-tuning\cite{mudrakarta2018k} & 80.12 & 87.54 & 91.32 & 70.31 & 78.45 & 81.54\\
Parallel Res. adapt \cite{rebuffi2018efficient} & {82.54} & {91.21} & {96.03} & 73.68 & {82.22} & 85.14\\
Series Res. adapt \cite{rebuffi2017learning} & 81.45 & 89.65 & 95.77 & 72.12 & 80.48 & 83.89\\
Piggyback\cite{mallya2018piggyback} & 81.59 & 89.62 & 94.77 & 71.33 & 79.91 & 83.45\\
TinyTL\cite{cai2020tinytl} & 82.34 & 90.23 & 94.63 & 71.39 & 80.44 & 83.80 \\\midrule
Ours (\mldam) & {\emph{83.33}} & {\emph{91.50}} & {\emph{96.65}} & \emph{72.79} & \emph{82.20} & \emph{85.29} \\\bottomrule
\end{tabular}}
\label{tab:accuracy}
\end{table*}


\begin{table*}[ht]
\centering
\caption{Summary of the results (i.e., activation memory(MB), training time (s)\% and inference computation (GFlops)) of the proposed method and comparison with the baseline techniques on four datasets ( e.g., CUBS, Stanford Cars, Flowers and Sketches) on NVIDIA Jetson Nano GPU. Note that, the reported training time is the time of training one epoch with batchsize 4 on average.}
\label{tab:cost}
\vspace{-1em}
\scalebox{0.85}{
\begin{tabular}{@{}cccccccc@{}}
\toprule
\multicolumn{4}{c}{Dataset} & Flowers & CUBS &  Cars & Sketches \\ \midrule
Methods & Model param (MB) & Active. mem (MB) & Inference GFlops & \multicolumn{4}{c}{\textless{} --------- Training Time (s) ---------- \textgreater{}} \\ \midrule
Standard Fine-tuning & {91.27} & 343.76 & 4.15 & 686 & 1977 & 2676 & 5843 \\
BN Fine-tuning & {91.27} & {174.17} & 4.15 & {173} & {507} & {683} & {1300}  \\
Parallel Res. adapt  & 177.8 & 308.8 &  4.68 & 558 & 1741 & 1604 & 4669   \\
Series Res. adapt  & 178 & 309.55 & 4.68 & 570 & 1832 & 1690 & 4783  \\
Piggyback & 94.12 & 343.76 & 3.44 & 1061 & 3015 & 4327 & 9783   \\\midrule
Ours (\mldam) & \emph{98.64} & {\emph{10.49}}  & \emph{3.17} & \emph{308} & \emph{834} & \emph{1073} & \emph{2274}  \\\bottomrule
\end{tabular}}
\end{table*}

Additionally, we also evaluate our proposed method on the Visual Decathlon Challenge \cite{rebuffi2018efficient}. The challenge is designed to evaluate the performance of learning algorithms on images from ten visual domains. 
The score ($S$) is evaluated as: $S = \sum_{i=1}^{10} \alpha_i \{0, E_{imax} - E_i\}^2$; where $E_i$ is the best error on domain $D_i$, $E_{imax}$ is the error of a reasonable baseline method, and the co-efficient $\alpha_i$ is the 1000$(E_i^{max})^{-2}$.


Finally, to evaluate the training efficiency of \mldam, we run our algorithm in the NVIDIA Jetson Nano GPU, which has 4GB DRAM with 20W power supply.
We evaluate the training time on this edge GPU (i.e., constrained memory) to demonstrate the memory-efficient training through \mldam.


\begin{table*}[ht]
\centering
\caption{Summary of the results (i.e., test accuracy \%) on the Visual Decathlon Challenge dataset. Here \# par denotes the number of model parameters with respect to a ResNet-26 baseline model in \cite{rebuffi2017learning}.}
\vspace{-1em}
\scalebox{0.75}{
\begin{tabular}{@{}cccccccccccccc@{}}
\toprule
 Methods & Model mem (MB) & Activ. mem (MB). & ImNet & Airc. & C100 & DPed & DTD & GTSR & Flwr & OGlt & SVHN & UCF & Score \\\midrule
Scratch  & {22.29} & 1315 & 59.87 & 57.1 & 75.73 & 91.2 & 37.77 & 96.55 & 56.30 & 88.74 & 96.63 & 43.27 & 1625 \\
Fine-tuning  & {22.29} & 1315 & 60.32 & 61.87 & 82.12 & 92.82 & 55.53 & {99.42} & 81.41 & 89.12 & 96.55 & 51.2 & 3096  \\
Series Res. adapt  & 24.94 & 1963 & 60.32 & 61.87 & 81.22 & 93.88 & 57.13 & 99.27 & 81.67 & {89.62} & 96.57 & 50.12 & 3159 \\
Parallel Res. adapt  & 23.62 & 1405 & 60.32 & 64.21 & 81.92 & 94.73 & {58.83} & 99.38 & 84.68 & 89.21 & 96.54 & {50.94} & 3412 \\
Piggyback & {22.29} & 1315 & 57.69 & {65.29} &79.87  & {96.99} &57.45  & 97.27 & 79.09 & 87.63 & {97.24} & 47.48  & 2838 \\\midrule
Ours (\mldam) & \emph{25.36}  & {\emph{201.7}} & {\emph{62.74}} & \emph{64.58} & {\emph{82.82}} & \emph{96.85} & \emph{59.43} & \emph{99.44} & {\emph{88.62}} & \emph{89.73} & \emph{97.47} & \emph{51.29} & \emph{{3498}} \\\bottomrule
\end{tabular}}
\label{tab:challenge}
\end{table*}

\paragraph{Baseline Methods.}
In this work, we primarily compare our method with three different baseline methods:
\begin{itemize}
    \item \textbf{Fine-tuning-based method:} There are mainly two general fine-tuning strategies. The first baseline fine-tunes all the parameters of the pre-trained model on each new dataset\cite{yosinski2014transferable}. Alternatively, the second one only fine-tunes the batchnorm and last classifier layers\cite{mudrakarta2018k}. 
    \item \textbf{Adaptor-based method:} This baseline learns a residual adaptor for each convolution layer, while freezing the pre-trained weights except batchnorm layer. We compare with three different residual adaptor designs: series adaptor~\cite{rebuffi2017learning}, parallel adaptor~\cite{rebuffi2018efficient} and TinyTL~\cite{cai2020tiny}. Note that, TinyTL is reproduced by applying the lite residual adaptor without network architecture search.
    \item \textbf{Mask-based method:} We choose piggyback \cite{mallya2018piggyback}, a popular binary mask learning scheme that keeps the under-lying pre-trained weights fixed. It only trains the binary mask to learn a large number of filters on top of a fixed set of pre-trained weights. 
\end{itemize}


\subsection{Results and Analysis}

We first compare our algorithm's efficacy with baseline methods by evaluating the performance on the test dataset listed in Table \ref{tab:accuracy}. Next, we evaluate the efficiency in reducing the training cost after deploying the models in NVIDIA Jetson Nano GPU in Table \ref{tab:cost}. The detailed experiments configuration is relegated in appendix-A.

\paragraph{Accuracy Comparison.}
In this evaluation section, each baseline method and \mldam train a ResNet-50 model with pre-trained weights on ImageNet dataset. As shown in Table \ref{tab:accuracy}, our proposed method \mldam achieves the \emph{best} test accuracy in CUBS, Stanford Cars and Flowers dataset. As for WikiArt, standard fine-tuning outperforms all the other techniques. 
Since WikiArt has a smallest number of samples between training and testing dataset in comparison to the other datasets, it helps to mitigate the over-fitting issue of fine-tuning the entire model. Finally, most notably, \mldam achieves comparable accuracy in comparison to the best baseline technique Parallel Res. Adapter \cite{rebuffi2018efficient}, achieving fractionally improved test accuracy in CUBS, Stanford Cars, and Flowers dataset, but much smaller training time shown in later Table \ref{tab:cost}. In summary, the proposed \mldam method achieves improved or comparable test accuracy in comparison to all the baseline techniques on five evaluation datasets.
\vspace{-1em}
\paragraph{Training and Inference Cost Comparison}
In Table \ref{tab:cost}, we summarize our key contribution in reducing the training and inference cost of multi-domain learning. Note, those are evaluated in a real memory-limited NVIDIA Jetson Nano GPU. As shown in the Table \ref{tab:cost}, the proposed \mldam method increases the model size by only a small fraction in comparison to Standard/BN Fine-tuning \cite{mudrakarta2018k} and Piggyback \cite{mallya2018piggyback} method. But \mldam reduces the activation memory size by \emph{19-37 $\times$} in comparison to the baseline techniques. As stated before, Parallel Res. Adapter \cite{rebuffi2018efficient} has shown superior performance (i.e., higher test accuracy) across four dataset. But we demonstrate that \mldam reduces the activation memory size by \emph{34 $ \times$} when compared with Parallel Res. Adapter's activation memory size while maintaining a similar test accuracy. 

\begin{figure}[ht]
  \centering
  \includegraphics[width=1.0\linewidth]{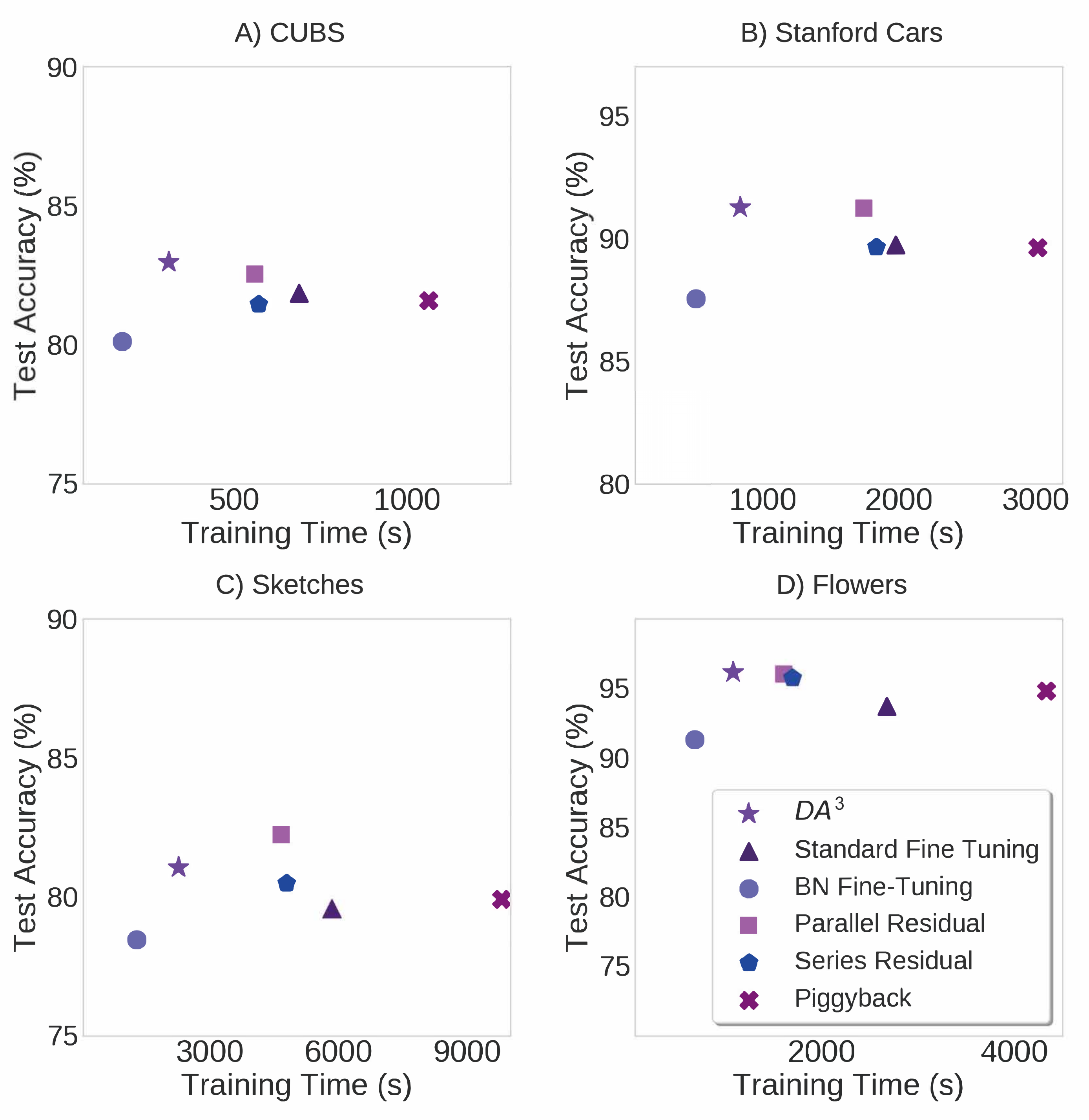}\\
 \vspace{-1em}
\caption{Trade-off between the Test Accuracy (\%) and Training Time (s) for four datasets A) CUBS, B) Stanford Cars , C) Sketches and D) Flowers.
}
\label{fig:trade}
\end{figure}

Apart from the reduction in memory cost, our proposed \mldam speeds-up the actual training time for on-device learning as well. As shown in Table \ref{tab:cost}, the training time reduces nearly by \emph{2 $\times$} in comparison to all the baseline techniques except for BN Fine-tuning \cite{mudrakarta2018k}. The faster training of BN Fine-tuning can be attributed to the presence of significantly less learnable parameters (less than 1MB), resulting in the worst accuracy performance in Table \ref{tab:accuracy}. Nevertheless, our method still outperforms BN based Fine-tuning in terms of both reduced activation memory size (i.e., 19 $\times$) and improved test accuracy across four datasets (e.g., CUBS, Stanford Cars, Flowers and Sketches). To summarize, in Fig. \ref{fig:trade}, we show that \mldam reduces training cost (i.e., time) in comparison to all the baseline methods (except BN fine-tuning); while maintaining on-par or improved test accuracy compared with the best (i.e., highest test accuracy \%) baseline method (i.e., Parallel Residual \cite{rebuffi2018efficient}).

Moreover, we also summarize the averaged inference computation cost on the five dataset in Fig. \ref{tab:accuracy}. Benefit from the spatial adaptor design, \mldam further achieves $1.30\times$, $1.47\times$ and $1.08\times$ inference computation cost reduction compared with fine-tuning-based, adaptor-based and mask-based method respectively.

\vspace{-1em}

\subsubsection{Visual Decathlon Challenge}

In Table~\ref{tab:challenge}, we show the effectiveness of our learning scheme on all the ten datasets of Visual Decathlon Challenge on ResNet-26. Note that, for this experiment, we plug an additive attention adaptor to each convolution layer. As reported in Table \ref{tab:challenge}, \mldam achieves $\sim$ 3 \% accuracy gain on ImageNet and  $\sim$ 4 \% accuracy gain on Flower dataset in comparison to the baseline methods. Moreover, it achieves the best $S$ score (3498) out of all the previous techniques demonstrating the effectiveness of our method in adapting to multi-domain tasks. Finally, it can also reduce the activation memory storage overhead during training by \emph{7-11 $\times$} in comparison to other methods; thus emerging as an ideal candidate for on-device learning purposes.

\subsubsection{Ablation Study of Additive Attention Adaptor}

We study the effectiveness of each component in the proposed additive attention adaptor on ImageNet-to-Sketch dataset setting. As shown in Table.\ref{tab:ablation}, we consider four different combinations to perform this ablation study: 1) Only updating bias (Only bias); 2) Only updating the basic adaptor module (Only Basic adap.); 3) Jointly updating the bias and spatial adaptor (Bias + Basic adap); 4) Jointly updating the proposed additive attention adaptor with bias.  
First, only bias has the worst accuracy, demonstrating the limited learning capacity using only a few bias parameters, supporting our initial hypothesis of adding the attention adapter to improve learning capacity. As a result, after adding the spatial adaptor, we observe a clear accuracy gain. Furthermore, jointly updating bias and spatial adaptor could improve accuracy even further. In the end, we introduce our proposed \mldam utilizing the channel attention module which connects the spatial adaptor in parallel to achieve the best performance. As \mldam succeeds in maintaining a reasonable test accuracy while drastically reducing the training overhead (as shown in Table \ref{tab:cost} \& Fig. \ref{fig:trade}).


\vspace{-0.5em}
\begin{table}[ht]
\centering
\caption{The ablation study on the proposed method}
\vspace{-1em}
\scalebox{0.8}{
\begin{tabular}{@{}cccccc@{}}
\toprule
Method & CUBS & Cars & Flowers & WikiArt & Sketch \\\midrule
Only bias & 74.53 & 83.85 & 87.30 & 68.73 & 71.93 \\
Only Basic adap. & 82.01 & 89.03 & 95.03 & 71.33 & 80.42 \\
Bias + Basic adap. & 82.15 & 89.73 & 95.56 & 71.88 & 80.70 \\\midrule
Proposed \mldam  & {\emph{83.33}} & {\emph{91.50}} & {\emph{96.65}} & \emph{72.79} & \emph{81.20}\\\bottomrule
\end{tabular}}
\label{tab:ablation}
\end{table}

\vspace{-1em}
\section{Conclusion}
We propose \mldam 
for memory-efficient on-device multi-domain learning, which is designed to eliminate the storage of intermediate activation feature maps. We design a method equipped with a novel additive attention adaptor to adapt original model to a new domain accurately. Such method not only reduces the training cost (e.g., time and memory) significantly, but also enables fast inference. Extensive experiments on domain adaption datasets consistently show the effectiveness and memory-efficiency of \mldam, paving a new way for memory-efficient on-device multi-domain learning.

\clearpage
\small{
\bibliography{egbib.bib}
}

\clearpage
\appendix


\end{document}


\subsection{A.Training Configuration.}
\begin{figure}[h]
  \centering
  \includegraphics[width=0.6\linewidth]{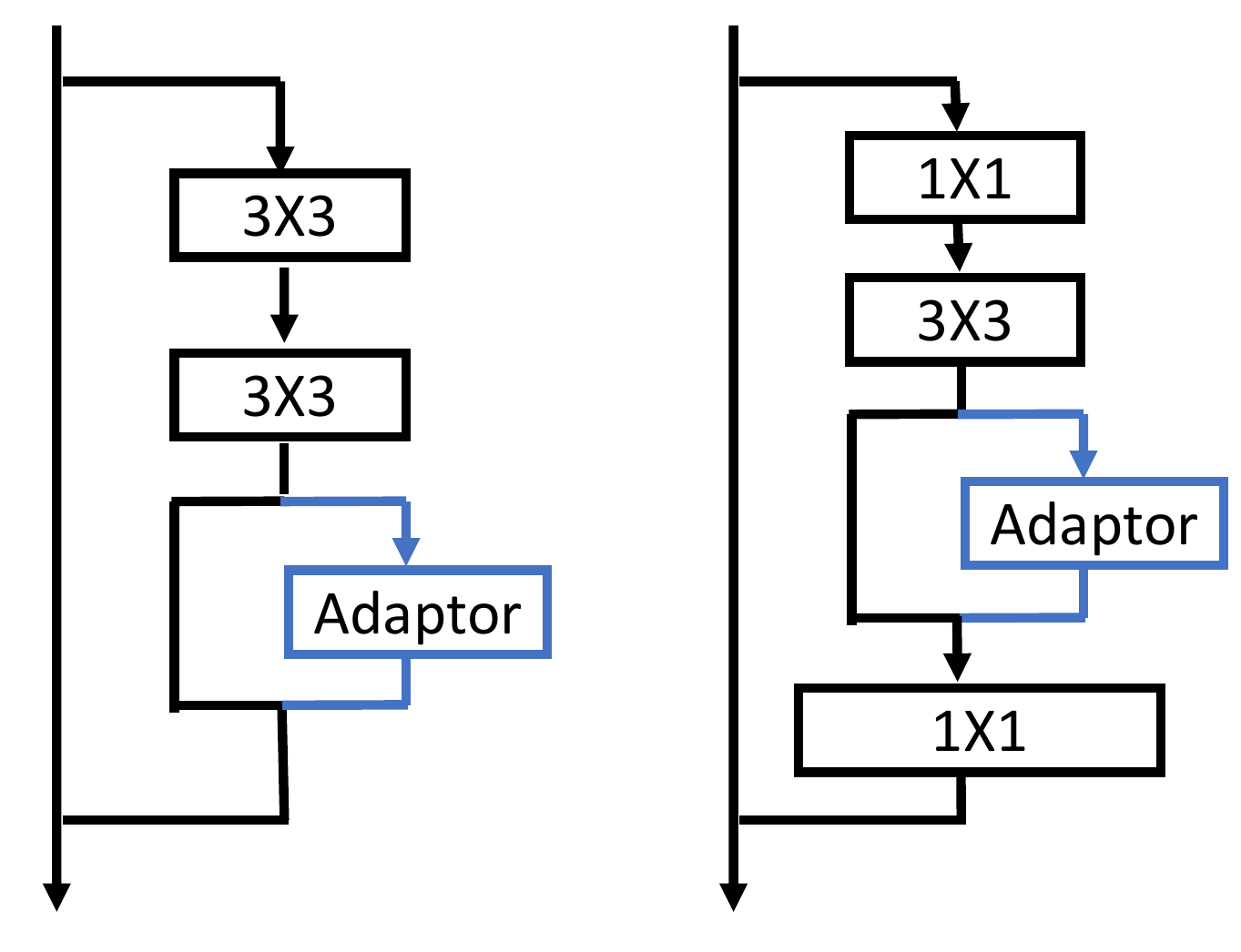}\\
\caption{Illustration of integrating the proposed \mldam in popular basic block and bottleneck block in ResNets. {Note, black indicates pre-trained backbone model, blue indicates added modules} 
}
\label{fig:ch_att}
\end{figure}

To demonstrate the efficiency of on-device training, we use Nvidia Jetson Nano GPU with 4-GB memory as our training platform. We evaluate the model performance using PyTorch as the simulation platform \footnote{https://forums.developer.nvidia.com/t/pytorch-for-jetson-version-1-7-0-now-available/72048}.Note that, the reported activation memory usage is calculated by our definition in Table.1, since PyTorch does not support explicit fine-grained memory management. In the training, for ResNet-50, we use Adam as the optimizer with cosine learning rate decay, an initial rate of 1e-3, and the number of iteration was set to 30. For ResNet-26 training on the challenge dataset, we use an SGD optimizer with an initial learning rate of 0.1. We schedule the learning rate decay at 40,80 and 100 epoch with a rate of 0.1. Again, as shown in Fig.\ref{fig:ch_att}, we use right configuration of \mldam for ResNet-50 and left configuration to train ResNet-26 model. 

\subsection{B.Learning the dynamic spatial gate}
First, we adopt a continuous logistic function:
\begin{equation}
    \sigma(\gG(\tH_s(\tA))) = \frac{1}{1+\textrm{exp}(-\beta\gG(\tH_s(\tA)))},
\label{eqt:sigmoid}
\end{equation}
where $\beta$ is a constant scaling factor. 
Note that the logistic function becomes closer to the hard thresholding function for higher $\beta$ values.

Then, to learn the binary mask, we leverage the Gumbel-Sigmoid trick, inspired by Gumbel-Softmax~\cite{jang2017categorical} that performs a differential sampling to approximate a categorical random variable. Since sigmoid can be viewed as a special two-class case of softmax, we define $p(\cdot)$ using the Gumbel-Sigmoid trick as:
\begin{equation}
    p(\gG(\tH_s(\tA))) = \frac{ \textrm{exp}((\textrm{log} \pi_0+g_0)/T)}{\textrm{exp}((\textrm{log}\pi_0+g_0)/T) + \textrm{exp}((g_1)/T)},
\label{eqt:soft_trick}
\end{equation}
where $\pi_{0}$ represents $\sigma(\vm^{r})$. $g_0$ and $g_1$ are samples from Gumbel distribution. The temperature $T$ is a hyper-parameter to adjust the range of input values, 
where choosing a larger value could avoid gradient vanishing during back-propagation. Note that the output of $\gG(\tH_s(\tA))$ becomes closer to a Bernoulli sample as $T$ is closer to 0. We can further simplify Eq.\ref{eqt:soft_trick} as:
\begin{equation}
    p(\gG(\tH_s(\tA))) = \frac{ 1}{1+\textrm{exp}(-(\textrm{log}\pi_0+g_0-g_1)/T)}
\label{eqt:gumb_simp}
\end{equation}

Benefiting from the differential property of Eq.\ref{eqt:sigmoid} and Eq.\cref{eqt:gumb_simp}, the real-value mask $\vm^{r}$ can be embedded with existing gradient based back-propagation training. To represent $p(\gG(\tH_s(\tA)))$ as binary format $\tG^{b}$, we use a hard threshold (i.e., 0.5) during forward-propagation of training. Because most values in the distribution of $p(\vm^{r})$ will move towards either 0 or 1 during training, generating the binary mask by $p(\gG(\tH_s(\tA)))$ could have more accurate decision, resulting in better accuracy.
